\documentclass[letterpaper, 10 pt, conference]{ieeeconf}  
\usepackage{amsmath,
amssymb, amsfonts}
\usepackage{algorithm}
\usepackage{algpseudocode}

\usepackage{graphicx}
\usepackage{textcomp}
\usepackage{xcolor}
\usepackage{adjustbox}
\usepackage{booktabs}
\usepackage{subcaption}
\usepackage{multirow}
\usepackage{svg}
\usepackage{float}
\usepackage{caption}
\usepackage{wrapfig}
\usepackage[normalem]{ulem}
\usepackage{arydshln}
\usepackage{caption}
\usepackage{ulem}

\newcommand{\model}{KCGG}
\newcommand{\motionpd}{MPD1}
\newcommand{\movementpd}{MPD2}
\newcommand{\ed}[1]{\textcolor{black}{#1}}
\newcommand{\edi}[1]{\textcolor{black}{#1}}
\newcommand{\edii}[1]{\textcolor{black}{#1}}
\newcommand{\ediii}[1]{\textcolor{black}{#1}}
\newcommand{\ediv}[1]{\textcolor{black}{#1}}
\newcommand{\edv}[1]{\textcolor{black}{#1}}
\newcommand{\cam}[1]{\textcolor{black}{#1}}
\newcommand{\rebut}[1]{\textcolor{black}{#1}}
\newcommand{\cut}[1]{\textcolor{red}{\sout{#1}}}
\renewcommand\cut[1]{} 
\newcommand{\cutt}[1]{\textcolor{red}{\sout{#1}}}
\renewcommand\cutt[1]{} 
\newcommand{\cuttt}[1]{\textcolor{red}{\sout{#1}}}
\renewcommand\cuttt[1]{} 
\newcommand{\cutttt}[1]{\textcolor{red}{\sout{#1}}}
\renewcommand\cutttt[1]{} 

\IEEEoverridecommandlockouts                              

\title{\LARGE \bf
Learning Diverse Robot Striking Motions with Diffusion Models and Kinematically Constrained Gradient Guidance
}

\author{Kin Man Lee$^{*1}$, Sean Ye$^{*1}$, Qingyu Xiao$^{1}$, Zixuan Wu$^{1}$, Zulfiqar Zaidi$^{1}$, David B. D'Ambrosio$^{2}$, \\ Pannag R. Sanketi$^{2}$ and Matthew C. Gombolay$^{1}$
\thanks{This work is supported in part by Naval Research Laboratory (NRL) N00173-21-1-G009, NSF CNS-2219755, and a gift award from Google.}%
\thanks{$^{*}$Equal contribution. $^{1}$Authors associated with the Institute of Robotics and Intelligent Machines (IRIM), Georgia Institute of Technology, Atlanta, GA, USA. $^{2}$Authors associated with Google DeepMind, Mountain View, CA, USA.}%
\thanks{Corresponding Author: Kin Man Lee,
 		{\tt\small klee863@gatech.edu}}
}
\begin{document}

\maketitle
\thispagestyle{empty}
\pagestyle{empty}

\begin{abstract}

Advances in robot learning have enabled robots to generate skills for a variety of tasks. Yet, robot learning is typically sample inefficient, struggles to learn from data sources exhibiting varied behaviors, and does not naturally incorporate constraints. These properties are critical for fast, agile tasks such as playing table tennis. Modern techniques for learning from demonstration \cut{such as Action Chunking with Transformers (ACT) }improve sample efficiency and scale to diverse data, but are rarely evaluated on agile tasks\cut{ where movement primitive (MP)-based approaches excel}. \ediv{In the case of reinforcement learning, achieving good performance requires training on high-fidelity simulators.} To overcome these limitations, we develop a novel diffusion modeling approach that is offline, constraint-guided, and expressive of diverse agile behaviors. The key to our approach is a kinematic constraint gradient guidance (KCGG) technique that computes gradients through both the forward kinematics of the robot arm and the diffusion model to direct the sampling process. KCGG minimizes the cost of violating constraints while simultaneously keeping the sampled trajectory in-distribution of the training data. \rebut{We demonstrate the effectiveness of our approach for time-critical robotic tasks by evaluating KCGG in two challenging domains: simulated air hockey and real table tennis. In simulated air hockey, we achieved a 25.4\% increase in block rate, while in table tennis, we achieved a 17.3\% increase in success rate compared to imitation learning baselines.}

\end{abstract}


\section{Introduction}

Controlling robot motion for agile athletic tasks, such as throwing, catching, or striking a ball, presents substantial challenges. These activities require algorithms for perception, prediction, and motor control, as well as robust hardware designed to handle dynamic actions. Moreover, these tasks demand high levels of precision, where minor errors in decision-making can lead to immediate failures. \rebut{Recent learning-based approaches for agile tasks are starting to rival hand-tuned control methods. Yet, specifying task execution within these learning-based frameworks often presents challenges. For reinforcement learning \edii{(RL)} methods, expert designers must define custom reward functions. Although inverse reinforcement learning~\cite{fu2017learning, chen2021learning} and adversarial imitation learning methods~\cite{ho2016generative} show potential, they frequently necessitate extensive hyperparameter tuning. In this paper, we approach the problem with imitation learning using a limited and diverse set of demonstrations.}

\edii{RL} and Learning from Demonstration (LfD) are the predominant robot learning methods used to solve agile tasks. \ediv{Recent works in RL for robotics have tried to improve sample-efficiency~\cite{luo2024serl} with advancements in sim-to-real transfer~\cite{buchler2022learning, piotr24}. However, these works rely on high fidelity simulators~\cite{haarnoja2018soft, dambrosio2024humanlevel, liu2022robot, peng2018sim,xie2020learning, koos2010crossing, boeing2012leveraging} and require experts to specify reward or cost functions which may not capture the true expert's intentions~\cite{dulac2021challenges}.} Similarly, LfD~\cite{billard2008robot, argall2009survey, ravichandar2020recent, chen2021learning, chen2023fast} techniques struggle to account for behavioral variance in multi-task settings. Recently, Transformer-based LfD methods such as ACT~\cite{zhao2023learning} have drastically improved the sample efficiency of deep learning-based LfD. Both \edii{RL} and LfD methods do not have the capability to enforce other constraints at test-time dynamically. RL policies assume a fixed MDP formulation and cannot adapt to new reward functions or task constraints. Deep learning-based LfD algorithms also typically do not have a formulation to incorporate novel kinematic or dynamic constraints at test time.

\begin{figure*}[t]
    \vspace{1mm}
    \centering
        \noindent\includegraphics[width=\textwidth]{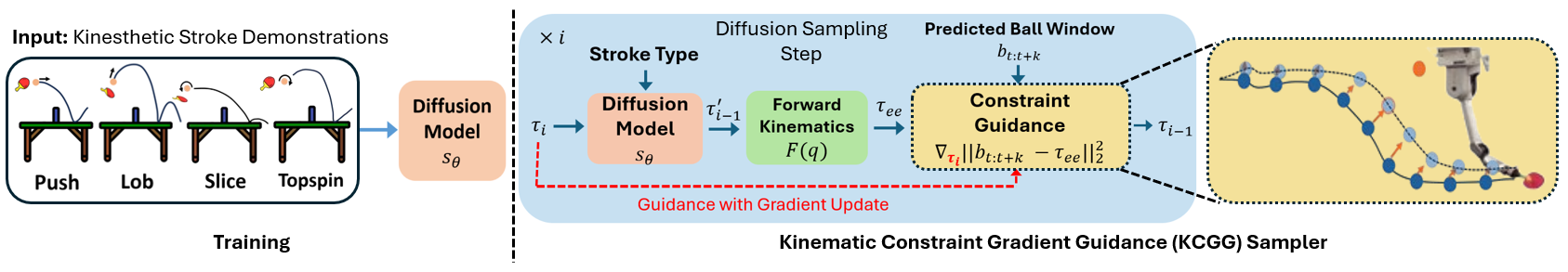}
        \caption{Overview of \model{}: \cutttt{Our diffusion model was trained on four kinesthetically demonstrated strokes.} \cam{At inference, \model{} computes the cost given the constraint and intermediate trajectory at each denoising step. The gradient with respect to the prior trajectory through the diffusion model is used to update the sample.}}
    \label{fig:block}
\vspace{-5mm}
\end{figure*}

In this paper, we develop a novel \rebut{imitation learning based} approach that enables robots to learn from small offline datasets to perform agile tasks in dynamical environments. We ground our work in an application to striking motions evaluated in established challenging domains for robotics: a virtual air hockey domain~\cite{liu2021efficient, liu2022robot} and a physical table tennis domain~\cite{gao2020robotic,muelling2010learning}. We show that our diffusion-based approach can learn to condition trajectories to meet environment constraints and \ediii{perform tasks difficult to specify}  (e.g. hitting a topspin, a high lob, etc) with just 15 kinesthetic demonstrations per stroke type\ediii{, which classical motion planning cannot achieve without explicit task heuristics}. The key to our approach is the novel formulation of a kinematic constraint gradient guidance (\model) technique that enables the robot to balance between constraints and adhering to the demonstration distribution.\cut{ The outcome of our work is a general framework for offline learning of high-speed robot behaviors that can afford task conditioning and diverse demonstration data as well as task-specific constraints on the desired robot trajectory.}

\noindent \textbf{Contributions:} Our key contributions are:
\begin{itemize}
    \cut{\item We demonstrate that our sampling method is capable of reproducing distinct, multimodal behavior captured by a diffusion model that is trained from a limited set of expert demonstrations ($<120$).}
    \item We demonstrate our sampling method can reproduce distinct, multimodal behavior with a diffusion model trained from limited expert demonstrations ($<120$).
    \item We propose \model, a novel state-of-the-art constraint sampling method for diffusion models. We show that \model{} outperforms baseline diffusion sampling methods in a simulated AirHockey domain. Our method achieves a 21.7\% increase in successful blocks.
    \cut{\item We show \model{} improves the ability for a real-world table tennis robot to return ping-pong balls successfully. Our method improves the success rate by 124.5\% over other diffusion baselines \cite{janner2022planning, carvalho23} and 17.3\% improvement over \ed{state of the art imitation learning} baselines \cite{zhao2023learning}, demonstrating our method's efficacy in agile robotic tasks.}
    \item We show \model{} improves a real-world table tennis robot's ability to return ping-pong balls. Our method improves success rate by 124.5\% over diffusion baselines \cite{janner2022planning, carvalho23} and 17.3\% over state-of-the-art imitation learning baselines \cite{zhao2023learning}, demonstrating efficacy in agile robotic tasks.
\end{itemize}

\section{Related Work}

We briefly review prior learning methods for robotic striking motions and key literature on diffusion models.

\paragraph{Learning Striking Motions} \edii{LfD}~\cite{billard2008robot, argall2009survey, ravichandar2020recent} aims to enable robots to acquire new skills by learning to imitate an expert. MP-based LfD has been explored extensively for agile striking motions in table tennis by learning movement primitives\cutt{ through kinesthetic teaching}. For example, Mixture of Movement Primitives (MoMP)~\cite{muelling2010learning, mulling2013learning} learns to select a dynamic movement primitive (DMP) from a library of pre-trained DMPs to generate joint trajectories for ball returns. Probabilistic movement primitives (ProMP)~\cite{paraschos2013probabilistic}, which model the variability in movements as trajectory distributions, were adapted\cutt{ for table tennis and lawn tennis} to generate striking motions that maximize the likelihood of ball return~\cite{gomezgonzalez2016using, gomezgonzalez2020adaptation, krishna2022utilizing, lee2023effect}. Movement primitive approaches are sample-efficient; however, generating diverse motions (e.g. slice vs. lob in table tennis) is difficult without a hierarchical approach\cutt{ to select the primitive to execute}. 


Recently, \edii{RL methods\cutttt{ that are utilizing heavy computational resources} are enabling systems to explore in task space without using demonstrated expert knowledge} \cite{gao2020robotic, tebbe2021sample, d2023robotic, abeyruwan2023agile, abeyruwan2023sim2real}. However, most of them require large training datasets~\cite{gao2020robotic, d2023robotic}, carefully shaped rewards~\cite{gao2020robotic}, or human behavior modeling~\cite{abeyruwan2023sim2real} which require a simulation. \ed{In contrast, we develop a fully offline method that requires few demonstrations and captures diverse agile behaviors in a single model, simplifying the training and deployment process onto real-world robots.}

\paragraph{Diffusion Models} Diffusion models~\cite{ho2020denoising, song2021scorebased} are a class of generative models\cutttt{ popularized in computer vision} that have recently shown promise in robotics for imitation learning~\cite{janner2022planning}, offline \edii{RL}~\cite{wang2023diffusion, ajay2023is}, and motion planning~\cite{janner2022planning, wu2024diffusion}. Compared with previous generative modeling approaches such as GANs~\cite{goodfellow2014generative}, diffusion models generally exhibit improved sample quality and diversity with a tradeoff in sampling time. In this work, we enhance the quality of samples generated by diffusion models while maintaining similar sampling time. This approach \cutttt{allows us to }produce\cam{s} higher-quality \cutttt{motion }samples without increasing computational cost, which is crucial for learning agile motions.

\rebut{Recent works using diffusion models for robotic manipulation, such as BESO~\cite{reuss2023goal} and \movementpd{}~\cite{scheikl2024movement}, demonstrated impressive performance and high success rates in challenging domains requiring millimeter-level precision. Our work, however, explores the additional complexity of solving dynamic tasks. Furthermore, while BESO~\cite{reuss2023goal} is designed for goal-conditioned tasks, the goal space in our domain is difficult to define. Consequently, our method does not rely on goal conditioning and instead adapts during sampling time.}

One advantage of diffusion models is the ability to guide samples towards certain objectives at sampling time. Several prior works in robotics have shown how to add constraints for motion planning \cite{carvalho23} and state estimation \cite{ye2023diffusion}. The computer vision community has also investigated using diffusion models to solve inverse problems~\cite{ongie2020deep, chung2023diffusion}, where the goal is to reconstruct samples from sparse measurements.\cut{ Examples of inverse problems in computer vision include inpainting~\cite{lugmayr2022repaint, saharia2022palette}, Computed Tomography (CT) reconstruction~\cite{mcleavy2021future, solomon2020noise}, and colorization \cite{Su2020Instance}. In our work,} We take inspiration from these techniques used in inverse problems to formulate a novel method for guiding diffusion samples towards kinematics constraints and evaluate this method in a difficult agile robotics domain. Our method is simple and enables the constraints to be better incorporated into the resulting diffusion samples. 

\begin{figure*}[t]
    \vspace{1mm}
    \centering
    \noindent\includegraphics[width=0.82\textwidth]{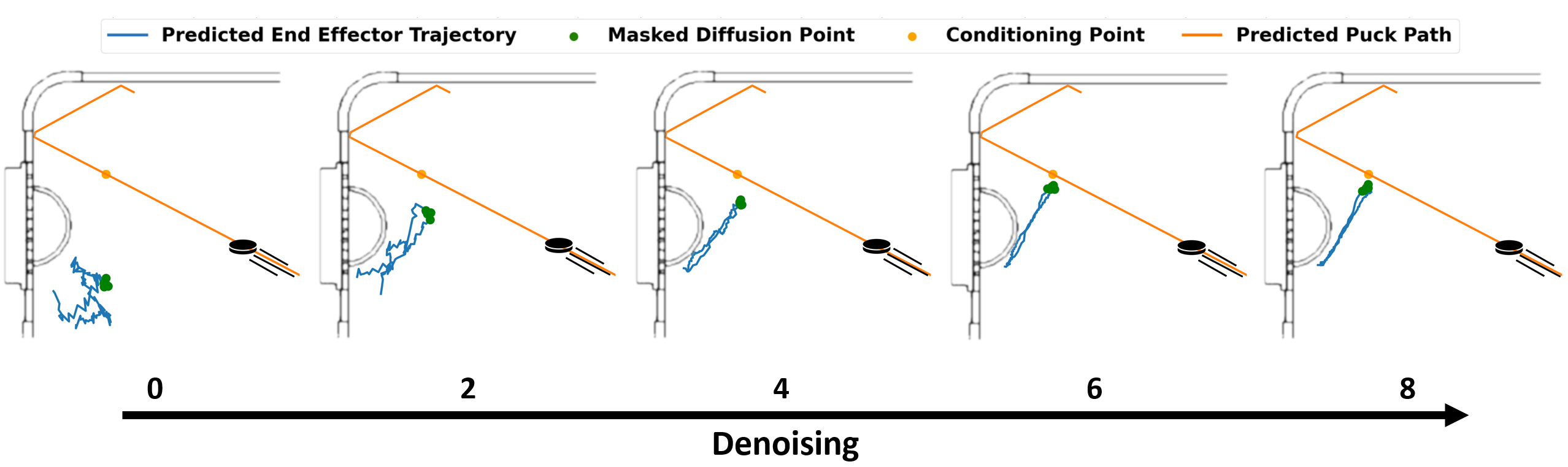}
    \caption{Visualization of Diffusion Sampling Process in Air-Hockey Defend: \model{} gradually shifts the conditioned points in the trajectory (shown in green) towards a predicted future puck position (shown as the orange point). }
    \vspace{-6mm}
\end{figure*}

\section{Problem Formulation}
\label{sec:problem_formulation}
The agile robot striking problem can be formulated as a motion planning task, where the objective is to determine a \rebut{trajectory} for the robot that adheres to a variety of constraints. We define $\tau = \{s_0, s_1, ... s_H\}$ as a sequence of states up to a time horizon\cam{,} $H$. Each state $s = [q, \dot{q}] \in \mathbb{R}^d$ is composed of the robot's joint positions\cam{,} $q$\cam{,} and joint velocities\cam{,} $\dot{q}$. The goal of motion planning is to identify a trajectory between initial and final states that satisfy a set of costs or constraints, $c_j(\tau)$. \cam{The objective is to minimize the total cost $\mathcal{J}(\tau) = \sum_j \lambda_j c_j(\tau)$, subject to kinematic or other constraints.} Other approaches formulate the motion-planning problem through probabilistic inference. Here, the optimality of a state within the trajectory is represented as $\mathcal{O}_t$ and the goal is to sample from a posterior distribution\cam{,} $p(\tau | \mathcal{O}_{1:t}) \sim p(\tau) p(\mathcal{O}_{1:t} | \tau)$\cutttt{, turning the motion planning problem into a conditional sampling problem} \cite{janner2022planning}. We use this formulation, where the diffusion model learns the prior, $p(\tau)$, and we use the guided sampling formulation \cite{ho2022classifier} to conditionally sample the model on our task constraints.\cut{ (See \ref{sec:derivation})}


A task constraint of interest for the agile striking problem is to ensure contact between the end-effector and the moving object, where $b \in \mathbb{R}^3$ is the object's position\cutt{ and $\dot{b} \in \mathbb{R}^3$ is the object's velocity} in Cartesian space. The forward kinematics function that maps the robot joint positions, $q$, to end-effector position, $x$, is denoted as $F\colon q \rightarrow x, q \in R^d, x \in R^3$. Our constraint minimizes the distance between the end-effector and object positions at a timestep, $t$, within a time window, $[t_s, t_e]$, where the robot can feasibly strike the object, i.e. $c(\tau) = \min_{t \in [t_s, t_e], q_t \in \tau}\|F(q_t) - b_t\|_2^2$. Our method is amenable to other types of constraints (e.g. for obstacle avoidance).

\section{Preliminaries: Score-Based Diffusion Models}
Diffusion models are a class of generative models that learn to generate samples through\cutt{the interplay of two processes: the} a forward noising process and \cutt{the}a reverse de-noising process. The forward noising process is commonly defined by: $d\tau_t = f(\tau_t, t) \, dt + \sigma(t) \, dw_t$, where $\tau$ is the state trajectory of the robot, $f(\tau_t, t)$ is the drift term and $\sigma(t)$ is the standard Brownian noise process. The forwards process is coupled with a backwards process: $d\tau_t = \left[ \mu(\tau_t, t) - \frac{1}{2} \sigma(t)^2 \nabla_{\tau_t} \log p_t(\tau_t) \right] dt$. In score-based diffusion models, the score function, $\nabla_{\tau_t} \log p_t(\tau_t, t)$, is typically parametrized by a neural network, $s_\theta(\tau, t)$. The model is trained with a score-matching loss: $\min_{\theta} \mathbb{E}_{p(\tau_t, t)} \left[ \left\| s_{\theta}(\tau_t, t) - \nabla_{\tau_t} \log p(\tau_t, t) \right\|_2^2 \right]$.






The formulation provided for score-matching diffusion above is of the continuous Stochastic Differential Equation (SDE) form, where the denoising timesteps are represented in continuous time. However, the SDE can be discretized to retrieve a discrete form that matches the Denoising Diffusion Probabilistic Models (DDPM) \cite{ho2020denoising} formulation commonly referred to as the Variance Preserving (VP) SDE~\cite{song2021scorebased}. In this formulation, the forward noising process is defined as: $\tau_i = \sqrt{\alpha_i} \tau_{i-1} + \sqrt{1 - \alpha_i} z, z \sim \mathcal{N}(0, I)$. \rebut{Following the convention from DDPM, $\beta_i$ represent the diagonal matrix of variances parameterizing the normal distribution, $\alpha_i = 1 - \beta_i$, and $\bar{\alpha}_i = \prod_{s=0}^i \alpha_s$. The variance parameter, $\beta_i$, is defined by the noise variance schedule, where we use the standard cosine schedule.} Complete noise is defined at $i=T$ while the data distribution is defined at $i=0$. \rebut{The relationship between the score-function, $s_\theta(\tau_i, i)$, and the model learned through DDPM's epsilon matching, $z_\theta(\tau_i, i)$, is $z_\theta(\tau_i, i) = - \sqrt{1 - \bar{\alpha}_i} s_\theta(\tau_i, i)$. For brevity, we omit $i$ and denote $s_\theta(\tau_i)$ as $s_\theta(\tau_i, i)$. }

\section{Methodology}
\label{sec:methodology}

In this section, we present the technical details of our novel sampling approach (Section \ref{sec:sampling}) for constrained sampling of diffusion models.

\subsection{Kinematic Constraint Gradient Guidance} \label{sec:sampling}
In this subsection, we outline the standard formulation for sampling from diffusion models without incorporating motion planning constraints. Then, we describe prior work for sampling with constraints and introduce our novel sampling approach. \cutttt{We refer to our method as Kinematic Constraint Gradient Guidance (\model).}

\paragraph{Standard Diffusion Model Sampling Procedure}
\cam{Each denoising step in a standard, unconstrained diffusion model follows Eq.~\ref{eq:tau}, where $s_\theta$ denotes the score function parameterized by the diffusion model and $z \sim \mathcal{N}(0, 1)$ is Gaussian noise parameterized by the denoising process. }

\begin{equation}
    \label{eq:tau}
    \tau_{i-1} = \frac{1}{\sqrt{\alpha_i}} \left(\tau_i + \left({1 - \alpha_i}\right)  s_\theta(\tau_i)\right)   + \sqrt{\sigma_i}z
\end{equation}

\paragraph{Constraint Guided Sampling} \label{sec:constraint-guided}
\cutttt{In order t}\cam{T}o generate trajectory samples conditionally with respect to the optimality condition, $p(\tau | \mathcal{O})$, prior works \cutttt{like Motion Planning Diffusion (\motionpd) \cite{carvalho23} and Diffuser \cite{janner2022planning}}choose to directly guide the noisy trajectories\cam{,} $\tau_{i-1}$\cam{, as follows in Eq.~\ref{eq:unconstrained}.}
\begin{equation}
\begin{aligned}
    \tau_{i-1}' &= \frac{1}{\sqrt{\alpha_i}} \left( \tau_i + \left({1 - \alpha_i}\right) s_\theta(\tau_i) \right),\\
    \tau_{i-1} &= \tau_{i-1}' - \mathbf{\nabla_{\tau_{i-1}'} c_j (\tau_{i-1}')} + \cam{\sqrt{\sigma_i}}z
\end{aligned}
\label{eq:unconstrained}
\end{equation}

\cam{However,}\cutttt{One problem} with this formulation,\cutttt{ is that} the constraint projects the intermediate diffusion samples out of distribution from the diffusion model's trained denoising paths \cite{choi2021ilvr}. We refer the reader to \cite{chung2022improving} for proofs on the coherence of these projection-based methods for diffusion model sampling. 



\paragraph{Kinematic Constraint Gradient Guidance}
\edii{\model{}} modifies the diffusion sampling procedure to Eq. \ref{eq:mcg}.
\begin{equation}
\begin{aligned}
    \tau_{i-1} &=  \tau_{i-1}' - \mathbf{\nabla_{\tau_{i}} c_j (\hat{\tau}_0)} + \cam{\sqrt{\sigma_i}}z
\end{aligned}
\label{eq:mcg}
\end{equation}

There are two modifications to prior work. 1) The gradient of constraints are computed for $\hat{\tau_0}$ rather than $\tau_{i-1}'$ and 2) The gradient is computed \textit{with respect to} $\tau_i$ rather than $\tau_{i-1}$.

By computing the gradient of the constraint for the predicted clean trajectory, $c_j(\hat{\tau_0})$, rather than a noisy trajectory, $c_j(\tau_{i-1}')$, the constraints can be better informed. Crucially, the cost functions, $c_j$, are not defined for noisy trajectories but rather noiseless trajectories. Therefore by using $\hat{\tau}_0$ we get a better gradient from the cost function. We can approximate $\hat{\tau_0}$ via $\tau_{i-1}$, \edv{as shown in Equation \ref{eq:relate}.}\cut{ Derivations in \ref{sec:derivation}.} 
\begin{equation}
    \label{eq:relate}
    \hat{\tau}_0 \approx \frac{1}{\sqrt{\overline{\alpha_i}}} \left(\tau_i + (1 - \overline{\alpha}_i) s_\theta(\tau_i, i)\right)
\end{equation}

Second, we compute the gradient with respect to $\tau_{i}$ rather than $\tau_{i-1}$. $\tau_i$ is the noisy trajectory \textit{before} the denoising step. \cutttt{The reason w}\cam{W}e choose $\tau_i$ rather than $\tau_{i-1}$ \cutttt{is that}\cam{because} we derive our procedure as sampling from the posterior distribution of $p(\tau|c)$. As we define our process through a score-based approach, we compute $\nabla_{\tau_i} \log p(\tau|c)$. Using Bayes rule, we therefore can compute the posterior with $\nabla_{\tau_i} \log p(\tau_i|c) = \nabla_{\tau_i} \log p(c|\tau_i) + \nabla_{\tau_i} \log p(\tau_i)$.\cut{ Further derivations are in \ref{sec:derivation}.}

\begin{figure}[!t]
    \begin{minipage}{\dimexpr\linewidth-4mm\relax}
        \begin{algorithm}[H]
        \caption{Batched Sampling with \model}
        \begin{algorithmic}[1]
        \Require Constraint function $c_j$
        \State $\boldsymbol{\tau}_T \gets$ sample \rebut{batch of trajectories} from $\mathcal{N}(0, I)$
        \label{line:noise}
        \For{all $i$ from $T$ to $1$} \label{line:for}
            \State $\hat{\mathbf{s}} \gets s_\theta(\boldsymbol{\tau}_i, i)$ \Comment{\rebut{Predict} score}
            \label{line:score}
            \State $\hat{\boldsymbol{\tau}}_0 \gets \frac{1}{\sqrt{\overline{\alpha}_i}} \left(\boldsymbol{\tau}_i + (1 - \overline{\alpha}_i)\hat{\mathbf{s}}\right)$ \Comment{Estimate $\hat{\boldsymbol{\tau}}_0$}
            \label{line:tau_zero}
            \State $\mathbf{z} \sim \mathcal{N}(0, I)$ \Comment{Sample Noise}
            \label{line:sample}
            \State $\boldsymbol{\tau}'_{i-1} \gets \sqrt{\frac{\alpha_i(1-\overline{\alpha}_{i-1})}{1-\overline{\alpha}_i}} \boldsymbol{\tau}_i + \sqrt{\frac{\overline{\alpha}_{i-1}\beta_i}{1-\overline{\alpha}_i}} \hat{\boldsymbol{\tau}}_0 + \tilde{\sigma}_i \mathbf{z}$
            \label{line:tau_prime}
            \State $\boldsymbol{\tau}_{i-1} \gets \boldsymbol{\tau}'_{i-1} - \nabla_{\boldsymbol{\tau}_i}(c_j(\hat{\boldsymbol{\tau}}_0))$ \Comment{\textbf{\model{} update}}
            \label{line:mcg}
        \EndFor
        \State $\tau_0^* \gets \arg\min(c_j(\boldsymbol{\tau}_0))$ \Comment{Batch filter}
        \label{line:select_best_tau}
        \State \Return $\tau_0^*$
        \end{algorithmic}
        \label{algo:sampling_kmcg}
        \end{algorithm}
    \end{minipage}
    \vspace{-4mm}
\end{figure}


\rebut{Given that $\hat{\tau_0}$ is a function of $\tau_i$, our approach takes the gradient not only through the constraint function}, $c_j$\cam{,} as prior methods did, but also through the diffusion model, $s_\theta$\cutttt{ itself}. Intuitively, by taking the gradient through both\cutttt{ the constraint function} $c_j$ and\cutttt{ the diffusion model} $s_\theta$, \cutttt{we modify }the entire trajectory\cam{,} $\tau_{i-1}$\cam{,}\cutttt{ to} adhere\cam{s} to the constraint rather than just a portion of the trajectory. \rebut{This whole-trajectory update is generally superior to updating small portions as it ensures the entire trajectory remains coherent and smooth.\cutttt{, with all timesteps adjusted in relation to each other. It allows for global optimization of the trajectory, potentially finding better solutions that satisfy the constraint while maintaining overall motion quality.} Furthermore, by updating the whole trajectory, we reduce the risk of getting stuck in local minima that may occur when only modifying small portions.}

Many motion planning task constraints, \cam{e.g. constraints on goal states}, only act on a \cutttt{few timesteps within}\cam{subset of} the trajectory. \cutttt{Similarly, our constraints only enforce the end-effector to be at the ball at a single timestep. }In \cutttt{traditional}\cam{prior} approaches, the function\cam{,} $\nabla_{\tau_{i-1}} c(\tau_{i-1})$ only updates \cutttt{a small portion}\cam{the constrained subset} of the predicted trajectory at each denoising step.\cutttt{, where all other points receive no updates as the gradient is 0. This limited update} \cam{However, this }can lead to suboptimal solutions, as it \cam{does not} consider the impact on the entire motion sequence. In contrast, our method's whole-trajectory update allows for more flexible and globally optimal solutions, even when constraints are localized to specific timesteps.

\subsection{Sampling Procedure} 
\label{sec:sampling_procedure}
We describe our guidance procedure in Algorithm \ref{algo:sampling_kmcg}. We predict a batch of trajectories, $\hat{\tau}_0$, with our network, $s_\theta(\tau_i, i)$ (\ed{L}ines \ref{line:score} - \ref{line:tau_zero}). An intermediate sample, $\tau_{i-1}'$, is computed using the standard DDPM procedure (\ed{L}ines \ref{line:sample} - \ref{line:tau_prime}). Then, the sample is updated \ediii{by KCGG} in \ed{L}ine \ref{line:mcg}. The function $F$ represents the forward kinematics chain for the robot arm.\cut{ The function is implemented in PyTorch such that we can automatically compute the gradients through both the forward kinematics and also the diffusion model to compute the constraint error.} We define our constraint function as 
$c(\tau) = \min_{t \in [t_s, t_e], q_t \in \tau}\|F(q_t) - b_t\|_2^2$, which minimizes the difference between the robot end-effector and ball prediction location at certain points in time defined by the window, $[t_s, t_e]$. Finally, we select a single best trajectory, $\tau_0^*$, based on the computed norm, $N$, from the final denoising step (Line \ref{line:select_best_tau}). This allows us to utilize the computational efficiency of batched operations on the GPU and select the trajectory that best matches the given end-effector constraint. We refer to this step as ``Batch Filter'' and \edi{ablate this feature}.

\begin{table}[t]
    \vspace{4mm}
    \centering
    \begin{tabular}{cc}
    \toprule
    Hyperparameter                       & Value   \\ 
    \midrule
    Learning Rate                        & 2e-4 \\
    Planning Horizon                     & 100   \\
    Batch Size                           & 32     \\
    Number of Training Diffusion Steps   & 100    \\
    Total Training Steps                 & 10e4    \\
    Optimizer                            & Adam     \\
    U-Net Convolutional Blocks           & (1, 4, 8)     \\
    \bottomrule
    \end{tabular}
    \caption{Architecture Hyperparameters in both domains}
    \label{tab:hyperparameters}
    \vspace{-6mm}
\end{table}


\section{Results and Discussion}
\label{sec:results}
In this section, we seek to address two primary questions: 
1) Do diffusion models have the capacity to generate diverse behaviors for executing the same task in multiple ways? 
2) Can \model{} facilitate trajectory generation from diffusion models to improve agile robotic tasks?

\subsection{\edi{Agile Robotic} Domains}
\label{sec:domains}
We evaluate on two domains: AirHockey\cutt{ Defend} and Table Tennis. In each domain, the object pose is estimated by a Kalman Filter~\cite{kalman1960new}. A predicted path of the object is obtained from a linear dynamics rollout in AirHockey\cutttt{,} and a ball dynamics model~\cite{chen2010dynamic} in Table Tennis. We form the time window, $[t_s, t_e]$, around a predetermined hit plane derived from the mean location of object hits in the collected demonstrations.


\subsubsection{Simulated Domain (AirHockey)}
We evaluate the air hockey simulation domain~\cite{piotr24} on two tasks, \textit{Hit} and \textit{Defend}. In \textit{Hit}, the objective is to score a goal with the desired shot type, \textit{Straight} or \textit{Angled}. A goal scored with the desired shot type is considered a \textit{Success}. In \textit{Defend}, the goal is to block an incoming puck that is initialized with random position and velocity. A \textit{Block} is defined as when the robot's end-effector successfully makes contact with the puck. Models are trained on 200 (100 per shot type) demonstrations in \textit{Hit} and 100 demonstrations in \textit{Defend}. Demonstrations are collected from a heuristic motion planner.

\subsubsection{Real-World Domain (Table Tennis)}
A 7-DOF Barrett WAM system from \cite{lee2023effect} with perception improvements~\cite{xiao2024multicamera} is used to return serves from a ball launcher. Training data is collected through kinesthetic teaching, with four types of shots: \textit{Push}, \textit{Lob}, \textit{Slice}, and \textit{Topspin}. Balls are launched to two locations on the table: \textit{Center} and \textit{Left}, with natural ball variance from the launcher. Models are trained on 120 demonstrations (15 per condition) and evaluated on three metrics: \textit{Success}, \textit{Hit}, and\textit{ Correct Stroke} rate. \textit{Success} is defined as a ball return within the spin and height distribution of the shot type's demonstrations. A \textit{Hit} is recorded when the ball contacts the end-effector, but fails the \textit{Success} criteria. Finally, \textit{Correct Stroke} is a sample that matches the desired shot type's joint trajectory distribution and striking location.

\begin{table}[t]
    \vspace{2mm}
    \centering
    \begin{tabular}{lrrr}
         Method&  Straight & Angle & Combined\\
         \toprule
         ProMP Unconditioned & - & - & 8.1\%\\
         Behavioral Cloning (BC-GMM)~\cite{mandlekar2021what} & 9.5\%  & 0.8\% & 5.2\%\\
         ACT~\cite{zhao2023learning} & 89.4\% & 65.2\% & 77.3\% \\
         Diffusion (Base) & \textbf{98.3\%} &  \textbf{82.4\%} & \textbf{90.4\%}\\
         
    \bottomrule
    \end{tabular}
    \caption{AirHockey Hit Results: Comparison of \textit{Success} rates on \textit{Straight} and \textit{Angled} shot types. }
    \label{tab:hit_results}
    \vspace{-1mm}
\end{table}



\subsection{Experiment Details and Baselines}

Due to the tight timing requirements of our domains, we use a mixture of both \cite{janner2022planning} and \cite{chi2023diffusionpolicy} as our baseline diffusion model, \cam{which we refer to as ``Diffusion (Base)"}. We utilize the 1D Convolutional Temporal U-Net from \cite{janner2022planning} due to its faster sampling speed over attention-based models. For conditioned models, we incorporate Feature-Wise Linear Modulation (FiLM) Layers \cite{perez2018film} to modulate the convolutional features within the U-Net, as suggested by \cite{chi2023diffusionpolicy}. Conditioned models use state-based observations of joint positions, stroke type, and the last two puck \cam{or last ten ball observations in the AirHockey and Table Tennis domains, respectively}. We follow the standard training procedure from DDPM~\cite{ho2020denoising} to train all diffusion baselines. Model hyperparameters used for all domains are shown in Table \ref{tab:hyperparameters}.

\edii{We benchmark against the following LfD baselines: Behavioral cloning with Gaussian Mixture Models (BC-GMM)} \cite{mandlekar2021what}, Probabilistic Movement Primitives (ProMP)~\cite{paraschos2013probabilistic}, and Action Chunking with Transformers (ACT)~\cite{zhao2023learning}. We also compare against four diffusion baselines: Diffusion (Base), Motion Planning Diffusion (\motionpd) \cite{carvalho23}, \rebut{BEbhavior generation with ScOre-based Diffusion Policies (BESO)~\cite{reuss2023goal} and Movement Primitive Diffusion (\movementpd)~\cite{scheikl2024movement}.}\cut{ See implementation details about the baselines in \ref{sec:appendix_baselines}.}




\begin{table}[t]
\centering
\resizebox{\linewidth}{!}{%
    \centering
    \begin{tabular}{@{}lrrrr@{}}
    \toprule
    Method & Block Rate $\uparrow$ & Time (s) & T & ms/step \\
    \midrule
    \textbf{Imitation Learning Baselines} \\
    Behavioral Cloning (BC-GMM)~\cite{mandlekar2021what} & 15.5\% & 0.001 & - & - \\
    ACT~\cite{zhao2023learning} & 79.4\% & 0.011 & - & - \\
    \midrule
    \textbf{Unconditioned Diffusion Models} \\
    Diffusion (Base) & 12.9\% & 0.159 & 20 & 7.95 \\
    Diffusion (Base) w/ Batch Filter & 62.6\% & 0.161 & 20 & 8.05 \\
    \motionpd~\cite{carvalho23} & 63.5\% & 0.190 & 16 & 11.88 \\
    \model{} (Ours) & \textbf{85.2\%} & 0.189 & 10 & 18.90 \\
    \midrule
    \textbf{Puck Conditioned Diffusion Models} \\
    Diffusion (Base) & 46.2\% & 0.162 & 20 & 8.10 \\
    Diffusion (Base) w/ Batch Filter & 46.6\% & 0.163 & 20 & 8.15 \\
    \motionpd~\cite{carvalho23} & 58.2\% & 0.191 & 16 & 11.94 \\
    \rebut{BESO~\cite{reuss2023goal}} & 57.4\% & 0.191 & 35 & 5.45\\
    \rebut{\movementpd~\cite{scheikl2024movement}} & 66.3\% & 0.192 & 21 & 9.14 \\
    \model{} (Ours) & 47.3\% & 0.188 & 10 & 18.80 \\
    \bottomrule
    \end{tabular}}
    \captionof{table}{AirHockey Defend Results: We compare our constraint sampling method against imitation learning and diffusion baselines. ``T'' is the number of diffusion timesteps.}
    \label{tab:defend_results}
    \vspace{-5mm}
\end{table}

\subsection{Simulation Results\edii{: AirHockey}}



In this section, we discuss results on the simulated AirHockey \edii{\textit{Hit} and }\textit{Defend} tasks. \edii{For each task, w}e test on 1000 initial start configurations. Our results for \edii{\textit{Hit} (Table~\ref{tab:hit_results}) show that Diffusion (Base) captures diverse motions best, achieving highest success rates for each shot type. Results for the \textit{Defend} task (Table \ref{tab:defend_results})}, show \model{} enables unconditioned diffusion models to outperform all baselines, increasing block rate by 18.9\% over the best baseline. \rebut{We use a fixed wall-clock time budget of 200 ms and tune sampling steps accordingly.} Interestingly, \model{} does not improve puck-conditioned models. We hypothesize that the conditioned model overfits the score function to puck observations.

We conduct an experiment on \textit{Defend} by using an obstacle avoidance constraint with \model{} instead, i.e. $c(\tau) = \min_{t \in [t_s, t_e], q_t \in \tau} \frac{1}{\|F(q_t) - b_t\|_2^2}$, and achieve a $2.9\%$ block rate, showing \model{} is compatible with other constraints. 

\begin{figure}[t]
    \vspace{1mm}
    \centering
    \includegraphics[width=0.9\linewidth, keepaspectratio]{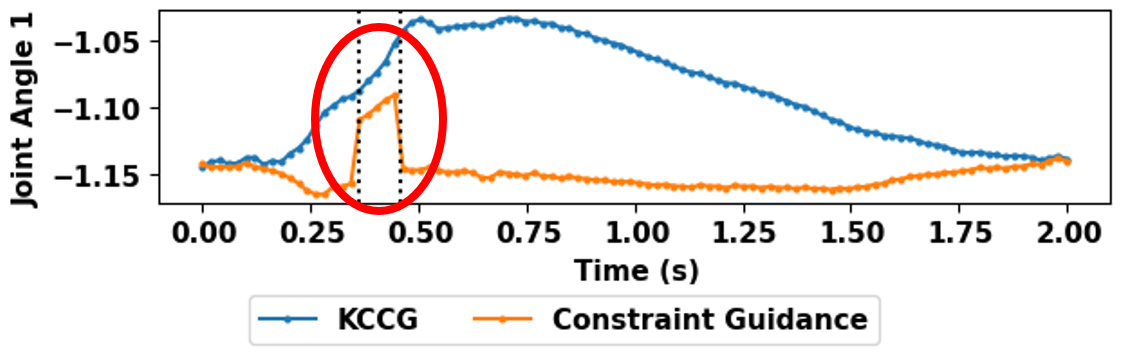}
    \caption{Comparison of Planned Trajectory: \model{} maintains a smooth joint angle profile in the constrained region (dotted lines) \ed{while constraint guidance introduces discontinuities.}}
    \vspace{-1mm}
    \label{fig:manifold_constraint}
\end{figure}

\begin{table}[t]
\centering
\resizebox{0.95\linewidth}{!}{%
    \centering
    \begin{tabular}{@{}lcc@{}}
         & Constraint wrt. $\tau_{i-1}$ & Constraint wrt. $\tau_0$ \\
         \toprule
         Gradient wrt. $\tau_{i-1}$ & 63.5\% (\motionpd{} \cite{carvalho23}) & 65.0\% \\
         Gradient wrt. $\tau_{i}$ & 70.0\% & \textbf{85.2\% (Ours)} \\
         \bottomrule
    \end{tabular}}
    \caption{Success Rates reported ablating for design choices in KCGG in the AirHockey Defend domain}
    \label{tab:constraint_ablation}
    \vspace{-5mm}
\end{table}




Figure \ref{fig:manifold_constraint} shows a comparison of \cutttt{the}a joint angle trajector\cutttt{ies}\cam{y} generated by \model{} and \cutttt{the original constraint guidance}\cam{\motionpd~\cite{carvalho23}}, revealing \cutttt{\model{} more effectively achieves the desired end-effector trajectory while adhering to the data distribution. T}\cam{that t}he trajectory produced by \cutttt{the original constraint guidance}\cam{\motionpd~\cite{carvalho23}} exhibits significant inconsistencies in the constrained region\cut{ (indicated by the black dotted lines)} \edi{whereas \model{} samples a smooth trajectory from the data distribution.}


\begin{figure}[b]
    \vspace{-4mm}
    \centering
    \includegraphics[width=0.8\linewidth, keepaspectratio]{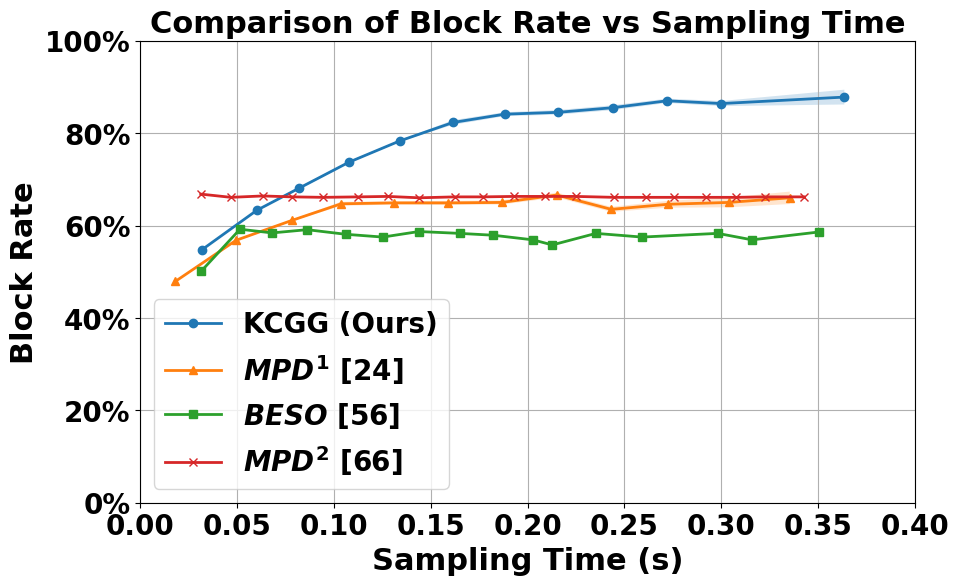}
    \caption{Sampling Speed vs Performance: We compare the Block Rate in AirHockey Defend across sampling times. \model{} scales to better performance with sampling time.}
    \label{fig:sampling_time}
\end{figure}

\begin{table*}[t]
   \vspace{3mm}
   \centering
   \resizebox{0.85\linewidth}{!}{
   \begin{tabular}{lccccccccccccc}
       \toprule
       & \multicolumn{3}{c}{Left Push} & \multicolumn{3}{c}{Left Lob} & \multicolumn{3}{c}{Left Slice} & \multicolumn{3}{c}{Left Topspin}\\
       \cmidrule(lr){2-4} \cmidrule(lr){5-7} \cmidrule(lr){8-10} \cmidrule(lr){11-13}
       Method & H & S & CS & H & S & CS & H & S & CS & H & S & CS \\
       \midrule
       ProMP~\cite{paraschos2013probabilistic} & 50\% & 20\% & - & 80\% & 30\% & - & 42\% & \textbf{26\%} & - & 66\% & \textbf{40\%} & -\\
       ACT~\cite{zhao2023learning} & 0\% & 0\% & 66\% & 64\% & 40\% & 88\% & \textbf{82\%} & 14\% & \textbf{100\%} & \textbf{70\%} & 24\% & 88\%\\
       Diffusion (Base) & 28\% & 8\% & 72\% & 38\% & 6\% & 46\% & 10\% & 4\% & 12\% & 14\% & 10\% & 22\%\\ 
       \motionpd~\cite{carvalho23} & 36\% & 22\% & \textbf{100\%} & 70\% & 16\% & \textbf{100\%} & 78\% & 8\% & \textbf{100\%} & 52\% & 16\% & \textbf{100\%}\\
       \rebut{BESO}~\cite{reuss2023goal} & 0\% & 0\% & 0\% & 0\% & 0\% & 0\% & 0\% & 0\% & 2\% & 0\% & 0\% & 0\%\\
       \rebut{BESO} w/ ball oracle~\cite{reuss2023goal} & 60\% & 46\% & \textbf{100\%} & 62\% & 28\% & \textbf{100\%} & 74\% & 2\% & \textbf{100\%} & 44\% & 10\% & \textbf{100\%}\\
       \rebut{\movementpd}~\cite{scheikl2024movement} & 72\% & 42\% & 94\% & 2\% & 2\% & 2\% & 4\% & 0\% & 6\% & 2\% & 2\% & 2\%\\
       \rebut{\movementpd} w/ ball oracle~\cite{scheikl2024movement} & 46\% & 20\% & \textbf{100\%} & 80\% & 32\% & \textbf{100\%} & 52\% & 0\% & \textbf{100\%} & 56\% & 34\% & \textbf{100\%}\\
       \model{} (Ours) & \textbf{76\%} & \textbf{50\%} & \textbf{100\%} & \textbf{96\%} & \textbf{60\%} & \textbf{100\%} & 64\% & \textbf{26\%} & \textbf{100\%} & 52\% & 26\% & \textbf{100\%}\\
       \midrule
       
       & \multicolumn{3}{c}{Center Push} & \multicolumn{3}{c}{Center Lob} & \multicolumn{3}{c}{Center Slice} & \multicolumn{3}{c}{Center Topspin}\\
       \cmidrule(lr){1-1} \cmidrule(lr){2-4} \cmidrule(lr){5-7} \cmidrule(lr){8-10} \cmidrule(lr){11-13}
       Method & H & S & CS & H & S & CS & H & S & CS & H & S & CS \\
       \midrule
       ProMP~\cite{paraschos2013probabilistic} & 84\% & \textbf{18\%} & - & 58\% & 24\% & - & 76\% & 0\% & - & \textbf{74\%} & 2\% & - \\
       ACT~\cite{zhao2023learning} & 56\% & 10\% & 90\% & 70\% & 26\% & 74\% & 28\% & 0\% & 34\% & 40\% & \textbf{22\%} & 62\% \\
       Diffusion (Base) & 42\% & 6\% & 42\% & 36\% & 10\% & 40\% & 38\% & 0\% & 98\% & 50\% & 16\% & 82\% \\
       \motionpd~\cite{carvalho23} & 66\% & 0\% & 94\% & 92\% & 18\% & \textbf{100\%} & \textbf{84\%} & 0\% & \textbf{100\%} & 72\% & 0\% & \textbf{100\%}\\
       \rebut{BESO}~\cite{reuss2023goal} & 2\% & 0\% & 2\% & 10\% & 4\% & 92\% & 0\% & 0\% & 0\% & 0\% & 0\% & 0\% \\
       \rebut{BESO} w/ ball oracle~\cite{reuss2023goal} & 94\% & 0\% & \textbf{100\%} & 92\% & 20\% & \textbf{100\%} & 40\% & 0\% & \textbf{100\%} & 44\% & 18\% & \textbf{100\%} \\
       \rebut{\movementpd}~\cite{scheikl2024movement} & 2\% & 0\% & 4\% & 44\% & 14\% & 86\% & 80\% & 0\% & \textbf{100\%} & 2\% & 0\% & 2\% \\
       \rebut{\movementpd} w/ ball oracle~\cite{scheikl2024movement} & 66\% & 0\% & \textbf{100\%} & 64\% & 12\% & \textbf{100\%} & 30\% & 0\% & \textbf{100\%} & \textbf{74\%} & 10\% & \textbf{100\%} \\
       \model{} (Ours) & \textbf{100\%} & 10\% & \textbf{100\%} & \textbf{100\%} & \textbf{32\%} & \textbf{100\%} & 72\% & 0\% & \textbf{100\%} & 70\% & \textbf{22\%} & 94\% \\
       \bottomrule
   \end{tabular}}
   \caption{Table Tennis Results: We compare the hit rates (H), success rates (S), and correct stroke rates (CS) for four different shot types across two different ball launcher configurations (left and center launches) with 50 trials per condition.}
   \label{tab:table_tennis_results}
    \vspace{-5mm}
\end{table*}

\subsection{Ablation Studies} 
\label{sec:defend_sampling}

\cam{Table~\ref{tab:constraint_ablation} reports ablation results for:} (1) Calculating the gradient with respect to $\tau_{i-1}$ or $\tau_0$ and (2) Computing the constraint function with $\tau_{i-1}$ or $\tau_0$. We find that taking the gradient with respect to $\tau_i$ brings a small improvement of 6.5\% over \motionpd{} and taking the constraint with respect to $\tau_0$ also provides a small 1.5\% improvement. With both, our \model{} method achieves large gains over the baselines.

We perform a sensitivity analysis against \rebut{other diffusion baselines} by varying the number of diffusion timesteps (Figure \ref{fig:sampling_time}).  \ed{Wall-clock time is evaluated as \rebut{sampling step time vary for each method.}} \ediii{All experiments
were run on a single Nvidia GTX 1080.} \ediii{Despite requiring more time per step, }\model{} \rebut{outperforms diffusion baselines across all sampling time budgets, except for \movementpd{} with sampling time budget less than $70$ ms}. As the number of diffusion timesteps increases, the performance of the baseline methods plateau. In contrast, \model{} scales, reaching a \ed{block} rate near 85\%, \rebut{while a method such as \movementpd{} does not, despite performing better with few sampling steps}. Additional denoising steps allows \model{} to better adhere the trajectory samples to both the training data distribution and task constraints. 



\subsection{Real-world Results: Table Tennis}
\cutttt{We report r}\cam{Results comparing \model{} to all other baselines are reported in Table \ref{tab:table_tennis_results}\cutttt{ with } (50 samples per task). }Based on AirHockey findings, \cutttt{ we condition}our models \cam{are conditioned} only on the stroke type to avoid over-parameterizing the score function.

In terms of task performance, \model{} shows an average 17.3\% improvement in success rate across all tasks, outperforming or matching the highest baseline success rates in six of eight tasks. Specifically, \model{} surpasses prior diffusion constraint guidance methods (\motionpd) with a notable 124.5\% improvement. We also note a modest average 4\% improvement in hit rate over the baselines\cutttt{, attributed mainly to high hit rates in the push and lob strokes}. 

\model{} excels in most strokes but has a lower hit rate in slice and topspin. This may be due to tighter timing requirements in these strokes, where less overlap between the ball and end-effector trajectories is present, posing challenges for \model{} in generating out-of-distribution trajectories (see Figure~\ref{fig:manifold_constraint}). While prior methods like \motionpd{} can adjust trajectories to meet hit points, they do not necessarily enhance success rates as they alter the stroke profile.


We evaluated \model's ability to express multimodal behaviors across four stroke types and two ball launch configurations. \model{} consistently demonstrated the desired behavior in each task, shown by high correct stroke rates. Conversely, ACT, despite conditioning on stroke style, struggled to generate the correct left/center strokes from ball observations, yielding a lower correct stroke rate. For BESO and \movementpd{}, we observed that conditioning on ball observation history caused overfitting and collapse into certain stroke modalities (e.g. BESO mostly samples Center Lob strokes), illustrating the importance of constraint guidance. We created stronger variants of these baselines by replacing ball history conditioning with ``ball oracle" conditioning, a categorical label indicating Left or Center launch distribution from the demonstration. This provides direct information on the ball launch location, which significantly improved BESO and \movementpd{} performance\cutttt{, removing mode collapse for a stronger comparison}. However, \model{} still outperformed these variants on average in hit and success rate. ProMP, which was not designed to capture multimodal behavior, was trained on data specific to each task, thus we exclude it from correct stroke rate evaluations. \model, however, effectively modeled trajectories across all conditions, accurately selecting the appropriate stroke based on the constraints.

\section{Limitations and Future Work} 
The primary drawback of our method is that we cannot generate trajectories that deviate significantly from the training data distribution. Because our approach relies on computing gradients through the trained score function, it produces samples that closely resemble behaviors in the training dataset. While this property ensures that the generated motions remain within the demonstrated range, it restricts the robot's adaptability to unseen scenarios. Future work could address this by using \model{} to warm-start RL.


\section{Conclusion}
\label{sec:conclusion}
Our work demonstrates \model{}'s effectiveness on small expert datasets in agile robotic tasks\cutttt{. Our algorithm is validated}\cam{, through evaluations} in both a simulated air hockey domain and on a real table tennis robot. \edv{Our real robot experiments show that we can learn diverse policies without access to a high-fidelity simulator}. These contributions address critical issues such as sample inefficiency and performance in dynamic settings, promising broader applications in real-world robotics. 

\newpage

\bibliographystyle{IEEEtran}
\bibliography{references.bib}

\clearpage

\end{document}